\documentclass[journal]{IEEEtran}

\usepackage{cite}
\usepackage{amsmath,amssymb,amsfonts}
\usepackage{algorithm}
\usepackage{graphicx}
\usepackage{makecell}
\usepackage{multirow}
\usepackage{color}
\usepackage{multirow}
\usepackage{xcolor}
\usepackage{xcolor}
\usepackage{algpseudocode}

\usepackage{textcomp}
\usepackage{wrapfig}
\usepackage[colorlinks,
            linkcolor=red,
            anchorcolor=blue,
            citecolor=green 
            ]{hyperref}

\begin{document}

\title{Cross-Scenario Deraining Adaptation with Unpaired Data: Superpixel Structural Priors and Multi-Stage Pseudo-Rain Synthesis}

\author{
Kangbo Zhao, Miaoxin Guan, Xiang Chen, Yukai Shi, Jinshan Pan
\thanks{ 

K. Zhao, M. Guan and Y. Shi are with School of Information Engineering, Guangdong University of Technology, Guangzhou, 510006, China (email: zhaokangbo329@gmail.com; guanmiaoxin1@mails.gdut.edu.cn; ykshi@gdut.edu.cn) . 

X. Chen and J. Pan are with the School of Computer Science and Engineering, Nanjing University of Science and Technology (email: chenxiang@njust.edu.cn; sdluran@gmail.com).

}

}
\maketitle

\begin{abstract}
Image deraining plays a pivotal role in low-level computer vision, serving as a prerequisite for robust outdoor surveillance and autonomous driving systems. While deep learning paradigms have achieved remarkable success in firmly aligned settings, they often suffer from severe performance degradation when generalized to unseen Out-of-Distribution (OOD) scenarios. This failure stems primarily from the significant domain discrepancy between synthetic training datasets and the complex physical dynamics of real-world rain. To address these challenges, this paper proposes a pioneering cross-scenario deraining adaptation framework. Diverging from conventional approaches, our method obviates the requirements for paired rainy observations in the target domain, leveraging exclusively rain-free background images. We design a Superpixel Generation (Sup-Gen) module to extract stable structural priors from the source domain using Simple Linear Iterative Clustering. Subsequently, a Resolution-adaptive Fusion strategy is introduced to align these source structures with target backgrounds through texture similarity, ensuring the synthesis of diverse and realistic pseudo-data. Finally, we implement a pseudo-label re-Synthesis mechanism that employs multi-stage noise generation to simulate realistic rain streaks. This framework functions as a versatile plug-and-play module capable of seamless integration into arbitrary deraining architectures. Extensive experiments on state-of-the-art models demonstrate that our approach yields remarkable PSNR gains of up to 32\% to 59\% in OOD domains while significantly accelerating training convergence. 
\end{abstract}


\section{Introduction}
\IEEEPARstart {I}mage deraining, which aims to remove rain streaks from rainy observations to recover the latent clean images, serves as a fundamental problem in low-level computer vision~\cite{chen2025towards,wang1909survey,yang2020single,yang2020single_tcsvt,weng2025dronesr}. Propelled by the rapid advancements in deep learning, deraining methodologies have witnessed remarkable improvements in both accuracy and robustness~\cite{yang2020single,zhang2023data,chen2021detail,wang2024idf}. Conventionally, deep 
 learning-based deraining is formulated as a supervised learning problem, necessitating extensive datasets of paired rainy inputs and ground-truth clean images for model training~\cite{yang2017deep,kang2011automatic,fu2017clearing,fu2017removing,zhang2022dynamic}. However, acquiring high-quality, large-scale paired datasets in real-world rainy scenarios poses a significant challenge, constrained by unpredictable weather conditions, hardware limitations, and prohibitive annotation costs. Consequently, most existing models resort to training on synthetic data, typically generated by superimposing simulated rain streaks onto clean backgrounds. 
\begin{figure*}[!t]
\centering
\includegraphics[width=0.9\linewidth]{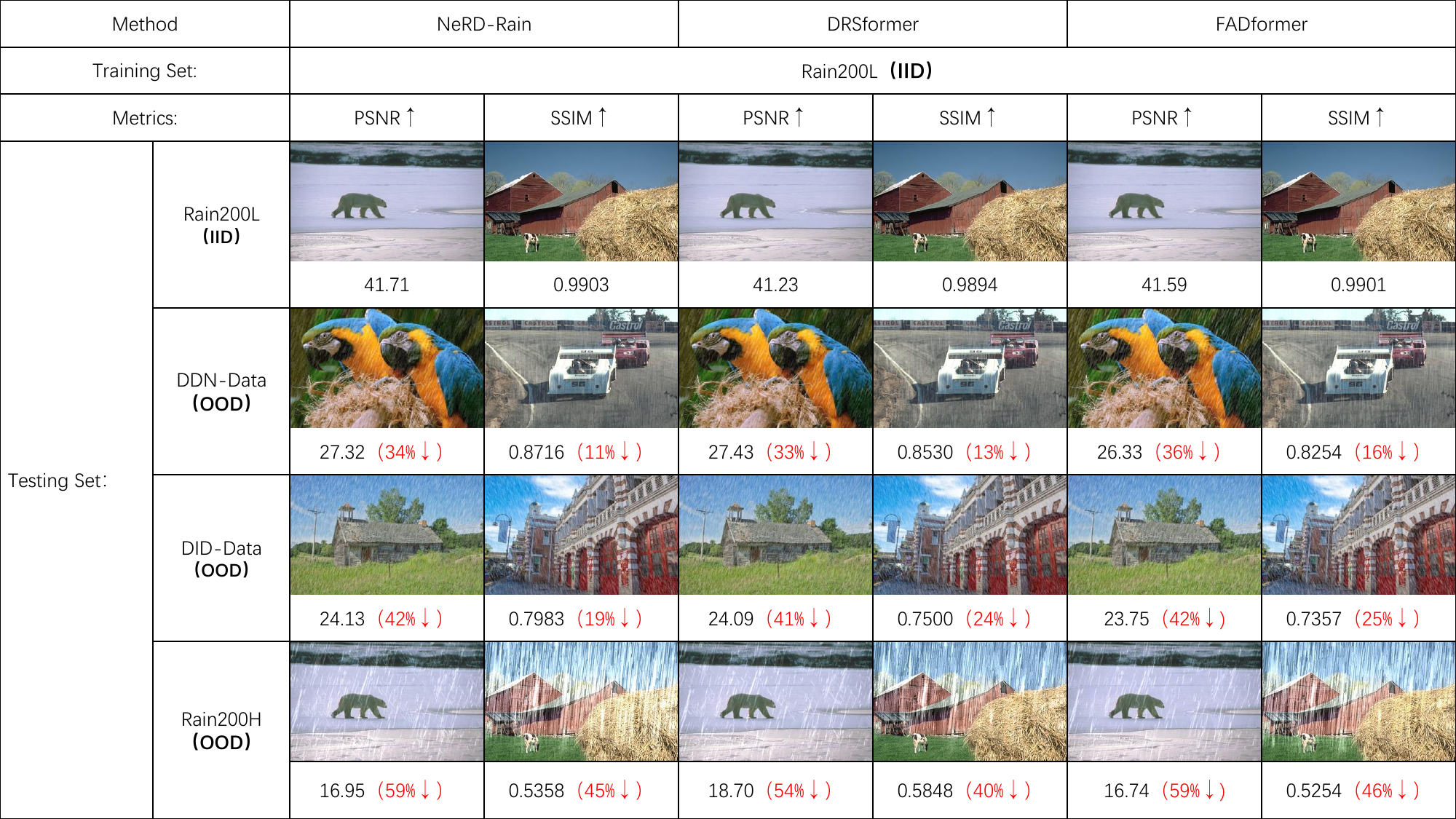}
\caption{The challenge of removing raindrop across various Out-of-distribution (OOD) scenarios. We first use the Rain200L dataset to train the NeRD-Rain, DRSformer, and FADformer methods respectively. In order to implement the evaluation on Out-of-distribution(OOD) scenarios, we use Rain200L as the source domain data, and Rain200H, DID, and DDN as the OOD domains for testing. Empirical results demonstrate that while the model achieves exemplary deraining performance on the source domain (Rain200L), the performance of state-of-the-art derain models deteriorates significantly when applied to unseen OOD domains. Thus, we call for a solid derain pipeline is required to better handle unknown scene conditions. }
\label{fig2}
\vspace{-3mm}
\end{figure*}

While this synthetic paradigm demonstrates efficacy in controlled settings and enables models to excel within the source domain, it often fails to faithfully replicate the complex physical dynamics of natural rain, including varying illumination, perspective occlusions, and haze interference~\cite{wang2021deep_tmm}. Consequently, these models suffer from substantial performance degradation when generalized to unseen target scenes, also known as Out-of-Distribution domains~\cite{yasarla2020syn2real,wei2019semi,li2022toward}. We regard datasets exhibiting distinct stylistic characteristics~\cite{yu2023single_tmm} as diverse scenarios and evaluate the performance of models pre-trained on Rain200L~\cite{yang2017deep}, including NeRD-Rain~\cite{chen2024bidirectional}, DRSformer~\cite{chen2023learning}, and FADformer~\cite{gao2024efficient}. While the models achieve impressive PSNR scores in the source domain, they suffer a precipitous performance decline ranging from 30\% to 60\% when applied to target domains such as Rain200H~\cite{yang2017deep}, DID-Data~\cite{zhang2018density}, and DDN-Data~\cite{fu2017removing}. These results underscore the severe domain shifts inherent in cross-domain scenarios. Empirically, the generated outputs frequently exhibit noticeable artifacts, including residual rain streaks, textural distortions, and even a loss of background fidelity. 

Consequently, generalizing image deraining to unseen scenarios has emerged as a pivotal research frontier demanding urgent attention in recent years~\cite{li2024revitalizing,fu2023continual}.
To thoroughly investigate the implications and challenges associated with this issue, we conceptualize "unseen scenes" as a series of datasets characterized by distinct stylistic divergences. Subsequently, we conduct a systematic evaluation of the generalization performance of deraining models trained on a single source dataset across multiple target datasets. As shown in Fig.~\ref{fig2}, experimental results demonstrate that the majority of contemporary deraining models suffer from a marked decline in performance during cross-dataset adaptation, frequently manifesting as residual rain streaks or over-deraining artifacts. These findings corroborate the ubiquity and severity of "cross-scenario domain discrepancies. "

Thus, domain discrepancies are pervasive in real-world applications, including intelligent security surveillance ~\cite{sultani2018real} and autonomous driving systems~\cite{kiran2021deep,yang2022semi}. Given the high heterogeneity of imaging devices and environmental conditions in these contexts, coupled with the scarcity of training data, there is an imperative need for a cross-scenario deraining methodology capable of autonomously adapting to stylistic variations. 

\begin{figure*}[!t]
\centering
\includegraphics[width=0.95\linewidth]{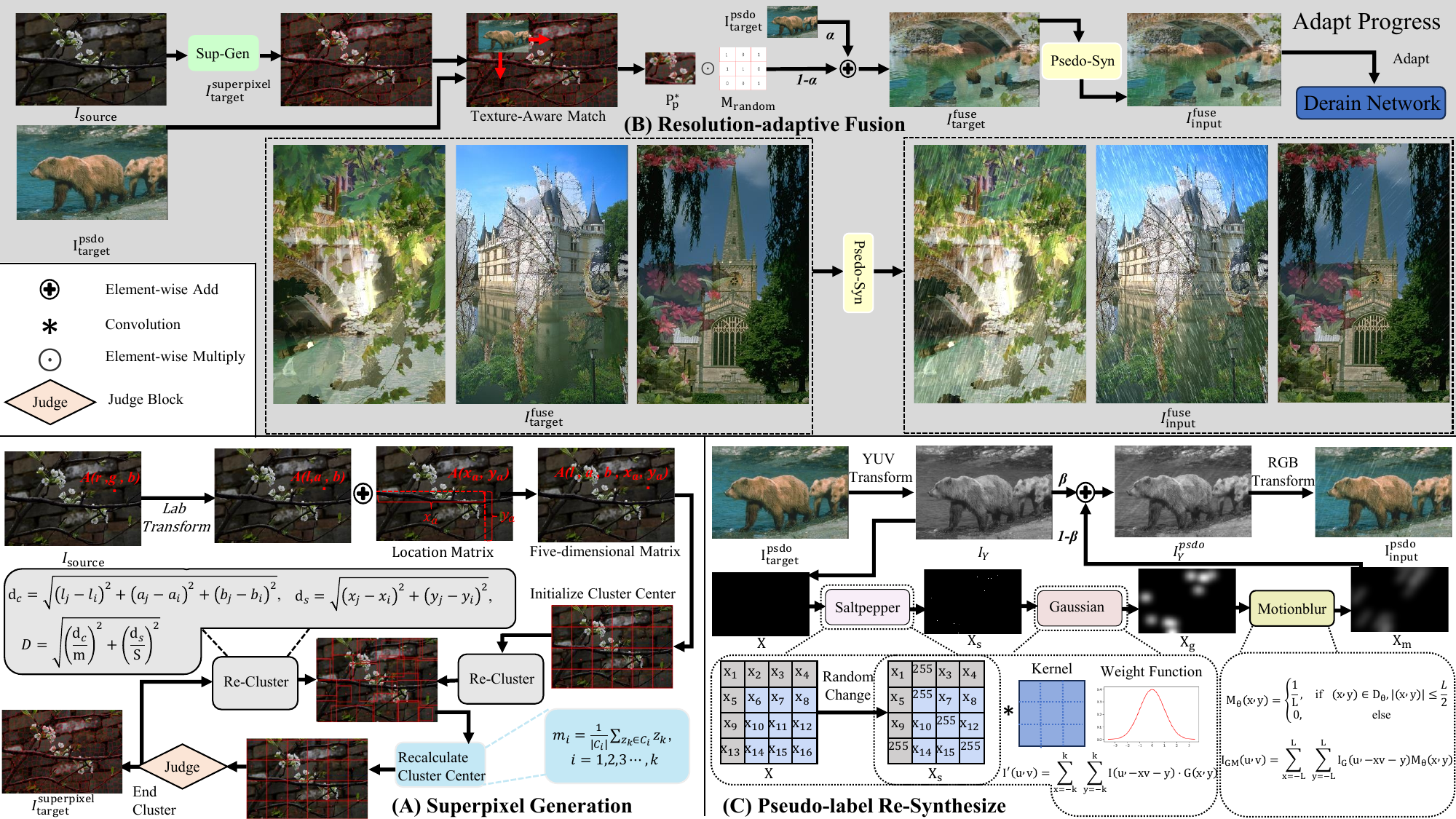}
\caption{An illustration of our proposed method: (A) Superpixel Generation: This module takes the source domain rain-free image $I_{source}$ as input. Utilizing a superpixel segmentation algorithm, it parses the image into a collection of superpixel blocks rich in structural and textural information, denoted as $I_{target}^{superpixel}$. (B) Resolution-adaptive Fusion: This module performs local matching between the extracted superpixel blocks $I_{target}^{superpixel}$ and the target domain’s pseudo rain-free image $I_{target}^{psdo}$. It locates optimal integration regions $\mathbf{P}_p^*$ based on semantic consistency and texture similarity(MSE). By employing a random mask matrix $M_{\text{random}}$ for proportional extraction and applying an $\alpha$-fusion strategy with a fusion coefficient of $\beta$, it generates an information-enhanced target image, $I_{target}^{fuse}$. (C) Pseudo-label Re-Synthesis: To synthesize high-fidelity pseudo rain streak layers, the module implements a three-stage degradation process comprising Salt-and-Pepper noise ($X_s$), Gaussian blurring ($X_g$), and motion blurring ($X_m$). This streak layer is superimposed onto the luminance channel of $I_{target}^{fuse}$ using $\alpha$ fusion to create the corresponding rainy image, $I_{input}^{fuse}$. With the synergistic operation of these three modules, a final set of pseudo-paired samples $(I_{input}^{fuse}, I_{target}^{fuse})$ is obtained, significantly enhancing the model's generalization toward out-of-distribution (OOD) domains. }
\label{fig1}
\vspace{-3mm}
\end{figure*}

To this end, this study presents a pioneering exploration of a cross-scenario deraining adaptation framework that operates independently of paired rainy and clean observations within the target domain. Requiring exclusively clean background images from the target dataset, the proposed method leverages a rain streak synthesis mechanism to progressively guide the model’s adaptation from the source to the target domain, thereby enabling it to capture the latent statistical characteristics intrinsic to the target scene. With the help of that, our model avoids the reliance on real paired data of the target domain, being more feasible and scalable in practical deployment.

In contrast to conventional Synthetic → Real or Real → Real adaptation paradigms, our proposed pseudo-label scene adaptation methodology aligns more closely with the intrinsic constraints and data characteristics of real-world applications. By effectively mitigating the domain discrepancy between training datasets and testing environments, this approach holds substantial potential for practical deployment toward non-aligned real-world scenarios. The primary contributions of this paper are summarized as follows: 
\begin{itemize}
\item[$\bullet$]We propose the first cross-scenario deraining adaptation framework that eliminates the need for rainy target observations. By training solely on rain-free target images, our method significantly enhances the feasibility of real-world deployment.
\item[$\bullet$]We introduce key information extraction and unsupervised augmentation mechanisms combining superpixel structural priors with texture matching. Utilizing Simple Linear Iterative Clustering (SLIC) for structural consistency and an MSE-based sliding window for alignment, we seamlessly blend source structures onto target backgrounds. This approach ensures natural fusion and improves the diversity of pseudo-data. 
\item[$\bullet$]Our method presents a plug-and-play flexibility toward existing deraining architectures without modification. Extensive evaluations on state-of-the-art models demonstrate that our approach accelerates convergence and significantly improves out-of-distribution generalization, achieving PSNR gains of 32\% to 59\%. 
\end{itemize}

\section{Related work}
\subsection{Image Deraining}
Image deraining aims to restore clean background scenes from rain-degraded observations, thereby enhancing the recognition accuracy of downstream computer vision tasks~\cite{yang2017deep,yi2021structure,wang2019spatial,ba2022not,shi2024nitedr}. Advancements in the field have driven a paradigm shift in deraining methodologies, evolving from conventional model-based approaches to data-driven deep learning techniques, and then to fusing attention mechanisms and Transformers~\cite{wang1909survey,yang2020single,zhang2023data}. 

Early iterations of image deraining methodologies were primarily predicated on physical models and hand-crafted priors~\cite{wang1909survey,yang2020single,kang2011automatic,li2016rain}. Notably, Zhang et al. introduced the Image Deraining Conditional Generative Adversarial Network (ID-CGAN)~\cite{zhang2019image}, which enhances both the visual fidelity and discriminative performance of the recovered images by incorporating generative adversarial mechanisms.

In recent years, deep learning methods have become the mainstream~\cite{yang2020single,yang2017deep,fu2017clearing,fu2017removing}. Convolutional Neural Networks (CNN) are widely applied to image deraining tasks~\cite{he2021dual,li2021single_scale,huo2021uncertainty,suh2022hierarchy,lin2021utilizing_tmm,wang2020dcsfn_tmm}. For example, the DID-MDN method~\cite{zhang2018density} significantly improves the deraining effect by jointly estimating rain density and deraining through a multi-stream densely connected convolutional neural network.

Propelled by the remarkable success of Transformers in computer vision~\cite{chen2021pre,han2022survey}, image deraining methodologies have increasingly incorporated Transformer-based architectures~\cite{xiao2022image}. A notable instance is Restormert~\cite{zamir2022restormer}, which mitigates computational complexity by calculating self-attention along the channel dimension. Additionally, TransWeather utilizes Transformer to handle multiple adverse weather conditions. 

Furthermore, the introduction of attention mechanisms has further improved the deraining effect~\cite{wang2019spatial,chen2021detail,fu2017removing,yu2022msaff_tmm}. For example, networks based on residual channel attention mechanisms~\cite{yi2021structure} recover clear images by accurately modeling the structure and details of the images~\cite{chen2024rethinking}.

To recapitulate, image deraining technology has undergone a significant evolution, progressing from conventional model-based approaches to data-driven deep learning paradigms, and subsequently to diverse architectures integrating Transformers and attention mechanisms~\cite{chen2025towards,wang1909survey,yang2020single,fu2021successive}, and more recently, Bayesian-based restoration frameworks~\cite{xiao2025bayesian}. Driven by the ongoing refinement of models and the proliferation of comprehensive datasets~\cite{chen2025towards,li2022toward}, image deraining technology holds great promise for achieving superior performance in real-world applications.

\subsection{Pseudo-Label Learning}
Pseudo-labeling represents a semi-supervised learning strategy widely adopted in deep learning~\cite{wei2019semi,huang2018multi,lu2024sirst,li2026ivan,lu2025rethinking}, designed to leverage unlabeled data to enhance the performance of the model. Sun \emph{et al.}~\cite{sun2025semi} use predictions of the model on unlabeled data as pseudo-labels, and then incorporate them to expand the training scale.

In the field of image deraining, the application of pseudo-label data augmentation is relatively rare, but the potential of it is gradually being focused on by researchers~\cite{liu2023embedding,yang2022semi}. Traditional image deraining methods mostly rely on manually annotated rainy images and clean image pairs~\cite{chen2025towards,li2022toward}, which presents problems such as high annotation costs and limited data volume in practical applications. To solve this problem, researchers attempt to utilize pseudo-label data augmentation strategies. 

For instance, Huang et al. introduced the Multi-pseudo Regularized Label (MpRL)~\cite{huang2018multi} technique. By assigning multiple pseudo-labels to generated imagery, this approach bolsters the capacity of the model to learn from unlabeled data, thereby yielding substantial performance gains in person re-identification tasks. 

Analogous strategies have also been investigated within the context of image deraining~\cite{yasarla2020syn2real,wei2020semi}. By employing techniques such as Generative Adversarial Networks (GANs)~\cite{zhu2019singe,wei2021deraincyclegan,zhang2019image,yasarla2020syn2real}, researchers generate pseudo-labeled data, which is subsequently co-trained with ground-truth labeled data to enhance the generalization capability of the model. In addition, combining self-supervised learning and pseudo-label strategies further improves the performance of the model on unlabeled data~\cite{jin2019unsupervised,liu2021unpaired,yu2021unsupervised,liu2023embedding} .

Nevertheless, the deployment of pseudo-label data augmentation strategies within image deraining remains in a nascent stage, confronted with critical hurdles including ensuring the quality of pseudo-labels and the mitigation of label noise~\cite{zhang2023data}.

In conclusion, pseudo-labeling strategies hold significant promise for image deraining applications. Future research can conduct in-depth exploration in aspects such as quality control of pseudo-labels and strategy optimization, to improve the performance and generalization ability of the model~\cite{chen2025towards}. 
\section{METHODOLOGY}
\subsection{Problem}

As Fig.~\ref{fig2} shows, we first apply the Rain200L dataset to train a few de-raining methods (e.g., NeRD-Rain, DRSformer, and FADformer). To test the results of state-of-the-art models on the source domain and the target domain respectively, we use Rain200L as the source domain data. And Rain200H, DID, and DDN are adopted as the Out-of-Distribution (OOD) domains for testing. Empirical results demonstrate that while the model achieves exemplary deraining performance on the source domain. Otherwise, the performance of the model deteriorates significantly when applied to unseen environments. 

\begin{figure}[t]
\centering
\includegraphics[width=0.95\linewidth]{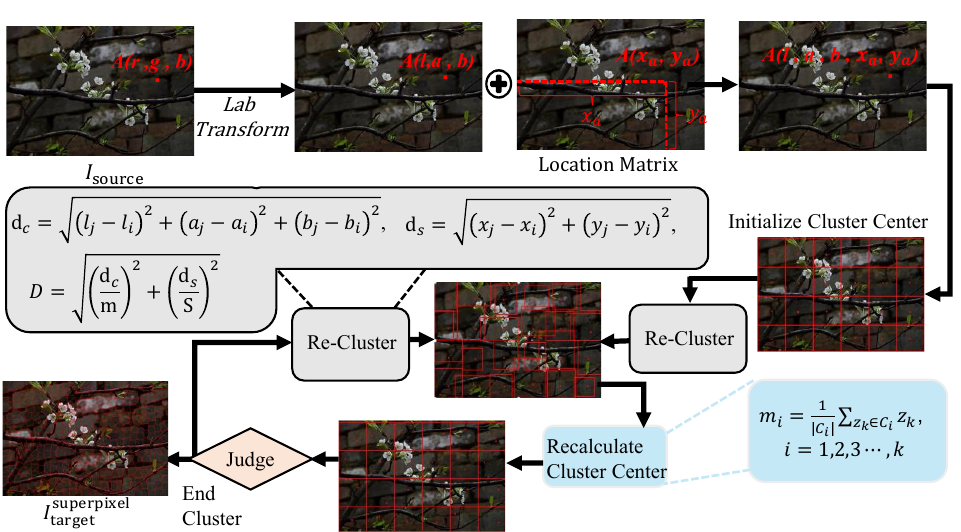}
\caption{The framework of Superpixel Generation, the process initiates by incorporating spatial positional information $(x,y)$ into the image data $(l,a,b)$. Subsequently, cluster centroids $(l_i,a_i,b_i,x_i,y_i)$ are initialized based on predefined parameters. }Clustering is then executed utilizing the SLIC ~\cite{6205760} algorithm, involving the re-calculation of updated cluster centroids.
\label{fig3}
\vspace{-3mm}
\end{figure}

To address the challenge of suboptimal deraining performance within unseen target domains, we propose across-domain adaptation module in Fig.~\ref{fig1}. It is mainly composed of three sub-modules: 
\begin{itemize}
\item[$\bullet$]Superpixel Generation (Sup-Gen) takes the source domain rain-free image $I_{source}$ as input, utilizing a superpixel segmentation algorithm to parse it into a set of superpixel patches $I_{target}^{superpixel}$ rich in structure and texture information$I_{target}^{superpixel}$. 

\item[$\bullet$]Resolution-adaptive Fusion performs local matching between $I_{target}^{superpixel}$ and the rain-free pseudo image of the target domain $I_{target}^{psdo}$. It locates the best integration region through correspondence relationships based on semantic consistency and texture similarity, utilizes a random mask matrix to extract superpixel blocks proportionally, and adopts an $\alpha$-blending strategy to synthesize the information-augmented target image $I_{target}^{fuse}$.

\item[$\bullet$] Pseudo-label Re-Synthesis (Psedo-Syn) generats high-fidelity pseudo-rain streak layers and superimposing them onto the two images respectively in the luminance channel in an $\alpha$-blending strategy the corresponding rainy images $I_{input}^{fuse}$ are synthesized. 
\end{itemize}

The synergistic integration of these three modules culminates in the generation of a set of pseudo-paired samples($I_{input}^{fuse}$,$I_{target}^{fuse}$). This outcome significantly bolsters the generalization and adaptive capabilities of the model within Out-of-Distribution (OOD) domains. 

\subsection{Superpixel Generation (Sup-Gen) }

To acquire high-quality local image patches that facilitate subsequent cross-domain fusion, this module employs the Simple Linear Iterative Clustering (SLIC) algorithm. This process partitions the rain-free source image $I_{\text{source}}$ into a collection of superpixel patches $I_{\text{target}}^{\text{superpixel}}$, which are characterized by structural compactness and feature consistency.

\begin{figure*}[t]
\centering
\includegraphics[width=1.0\linewidth]{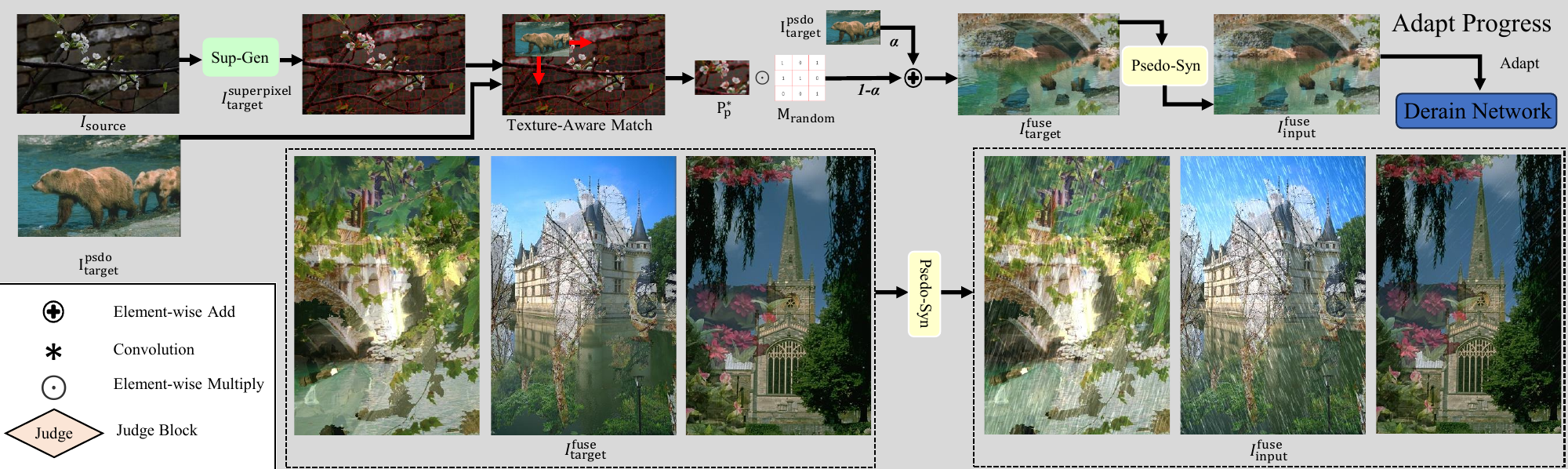}
\caption{The framework of Resolution-Adaptive Fusion. We propose a novel fusion paradigm driven by bidirectional region matching and adaptive fusion mechanisms. This approach simultaneously incorporates structural priors from the source domain while preserving the intrinsic background distribution of the target domain, thereby facilitating subsequent domain adaptation tasks. }
\label{fig4}
\vspace{-3mm}
\end{figure*}

As illustrated in Fig.~\ref{fig3}, the input image is initially transformed from the RGB to the CIELAB color space. Specifically, for each pixel $p_i$, we construct a 5D feature vector $V_i = [l_i, a_i, b_i, x_i, y_i]^T$, where $(l, a_i, b_i)$ represents the color coordinates in the CIELAB space and $(x_i, y_i)$ denotes the spatial position. To effectively associate these heterogeneous parameters while resolving the scale disparity between color and spatial domains, we employ a normalized distance metric:
\begin{equation}
\begin{aligned}
d_c &= \sqrt{(l_j - l_i)^2 + (a_j - a_i)^2 + (b_j - b_i)^2}, \\
d_s &= \sqrt{(x_j - x_i)^2 + (y_j - y_i)^2}, \\
D' &= \sqrt{ \left( \frac{d_c}{m} \right)^2 + \left( \frac{d_s}{S} \right)^2 }. 
\end{aligned}
\end{equation}

where $C_k$ and $P_i$ denote the feature vectors of the cluster center and the candidate pixel respectively. $d_c$ and $d_s$ are the Euclidean distances in the color and spatial domains, respectively. $S = \sqrt{N/k}$ is the initial grid interval (where $N$ is the pixel count and $k$ is the number of superpixels), and $m$ is the compactness parameter that balances boundary adherence and shape regularity. And we set $k$ to 50 here.

Regarding computational complexity, the integration of these 5D parameters does not impose a prohibitive burden. By constraining the search space for each cluster center to a $2S \times 2S$ local neighborhood, each pixel is compared only with a limited number of adjacent centers . Consequently, the complexity is reduced from the standard K-means $O(kNI)$ to a strictly linear complexity of $O(N)$. This ensures that our Sup-Gen module maintains exceptional efficiency even for high-resolution images across diverse scenarios.

The algorithm initializes cluster centers and iteratively performs pixel assignment and cluster center updates within a 2S$\times$2S region around each center until the residual converges. The following is the formula for recalculating cluster centers each time: 

\begin{equation}
\begin{gathered}
m_{i}=\frac{1}{|C_{i}|}\sum_{z_{k}\in C_{i}}z_{k}, 
i=1,2,3\cdots,k
\end{gathered}
\end{equation}

where $m_i$ denotes the updated cluster centroid, $z_k$ represents the feature vectors of constituent pixels within the cluster, and $C_i$ corresponds to the cardinality of the cluster region. This process ultimately yields the set of superpixel patches $I_{\text{target}}^{\text{superpixel}}$, serving as foundational building blocks rich in structural information for the subsequent fusion stage.

\subsection{Resolution-adaptive Fusion}

To incorporate structural information from the source domain, we propose a Resolution-Adaptive Fusion anchored in texture similarity and Mean Squared Error (MSE) matching. The inputs to this module comprise the set of source domain superpixel patches, denoted as $I_{\text{target}}^{\text{superpixel}} \in \mathbb{R}^{H_s \times W_s \times 3}$, and the rain-free pseudo-image of the target domain. Here, $H_s, W_s$ and $H_p, W_p$ the corresponding dimensions of the target image represent the spatial resolution of the superpixel patches and the target image, respectively.

This module aims to localise specific regions within the target image to facilitate Resolution-Adaptive Fusion with source domain superpixels. The procedural framework involves identifying the region that exhibits maximal texture similarity to the target image within the source domain superpixel patches. This is executed via a sliding window matching mechanism, where the alignment is optimised by minimizing the Mean Squared Error (MSE): 

\begin{equation}
\begin{aligned}
&(y^*, x^*) = \arg\min_{y,x} \frac{1}{3H_p W_p} \\ 
&\bigg\| I_{\text{target}}^{\text{superpixel}} 
\quad [y:y+H_p, x:x+W_p] - I_{\text{target}}^{\text{psdo}} \bigg\|_2^2
\end{aligned}
\end{equation}

where $I_{\text{target}}^{\text{superpixel}}[y:y+H_s, x:x+W_s]$ represents the candidate region extracted from the target image. After obtaining the optimal matching region $\mathbf{P}_s^*$, random sampling is performed on it, followed by adaptive fusion with $I_{\text{target}}^{\text{psdo}}$: 

\begin{equation}
I_{\text{target}}^{\text{fuse}} = I_{\text{target}}^{\text{psdo}} \cdot \alpha + \mathbf{P}_s^* \cdot M_{\text{random}} \cdot (1 - \alpha)
\end{equation}

where $M_{\text{random}}$ is a random mask matrix controlling the sampling position of superpixel blocks, and $\alpha$ is a hyperparameter for balancing the fusion weight. $I_{\text{target}}^{\text{fuse}}$ is the pseudo data after fusion. If we discover that $I_{\text{target}}^{\text{superpixel}}$ does not meet the fusion conditions, we directly match a region of $I_{\text{target}}^{\text{psdo}}$ to perform replacement transfer:

\begin{equation}
\mathbf{P}_p^{\text{fuse}} = \mathbf{P}_p^* \cdot \alpha + I_{\text{target}}^{\text{superpixel}} \cdot M_{\text{random}} \cdot (1 - \alpha)
\label{my_formula}
\end{equation}
\begin{equation}
I_{target}^{fuse}=I_{target}^{psdo}-\mathbf{P}_{p}^{*}+\mathbf{P}_{p}^{fuse}
\end{equation}

where $\mathbf{P}_p^*$ is the optimal matching region obtained in $I_{\text{target}}^{\text{psdo}}$. Through the application of bidirectional region matching and adaptive fusion mechanisms, this module incorporates structural priors from the source domain while simultaneously preserving the background distribution intrinsic to the target domain. As evidenced in Fig.~\ref{fig4}, the imagery processed by this module effectively integrates structural information from the source domain while maintaining the integrity of the target scene content. Consequently, the resultant fused pseudo-data exhibits enhanced diversity and stochasticity, supplying high-quality training samples for the subsequent domain adaptation of the deraining model. 

\begin{figure}[t]
\centering
\includegraphics[width=0.95\linewidth]{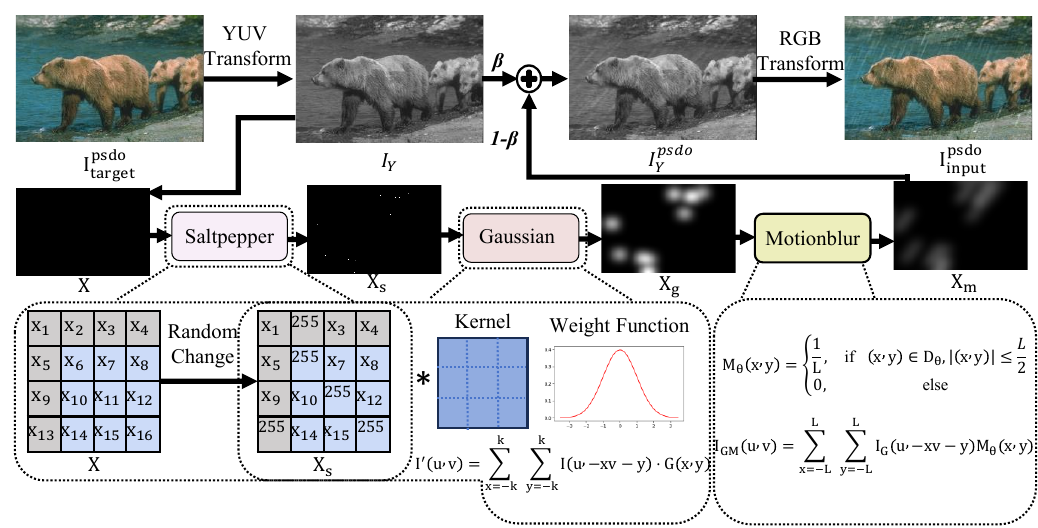}
\caption{Pseudo-label Re-Synthesis introduces a pseudo-rain streak generation methodology grounded in multi-stage noise synthesis. By sequentially executing noise initialization, morphological transformation, and luminance channel fusion within the YUV color space, the proposed method achieves the realistic simulation of rain effects on rain-free images within the target domain.}
\label{fig5}
\vspace{-3mm}
\end{figure}

\subsection{Pseudo-label Re-Synthesis (Psedo-Syn)}

As shown in Fig.~\ref{fig5}, specifically, given the rain-free input imagery from the preceding module (i.e., comprising $I_{\text{target}}^{\text{psdo}}$ and $I_{\text{target}}^{\text{fuse}}$ ) the process initiates by transforming these inputs into the YUV color space and extracting the luminance component $I_Y$. Subsequently, we generate rain streaks through three stages: First, create a rain streak mask $X$ and inject salt-and-pepper noise with a density of $p$ to obtain the point-like noise base $X_s$. Subsequently, apply a $k \times k$ Gaussian kernel (standard deviation $\sigma_g$) to perform blurring processing, forming the patch-like noise $X_g$ with optical gradient characteristics. Finally, convert the patch-like noise into linear rain streaks through motion blur, generating the final rain streak mask $X_m$:

\begin{figure*}[t]
\centering
\includegraphics[width=0.9\linewidth]{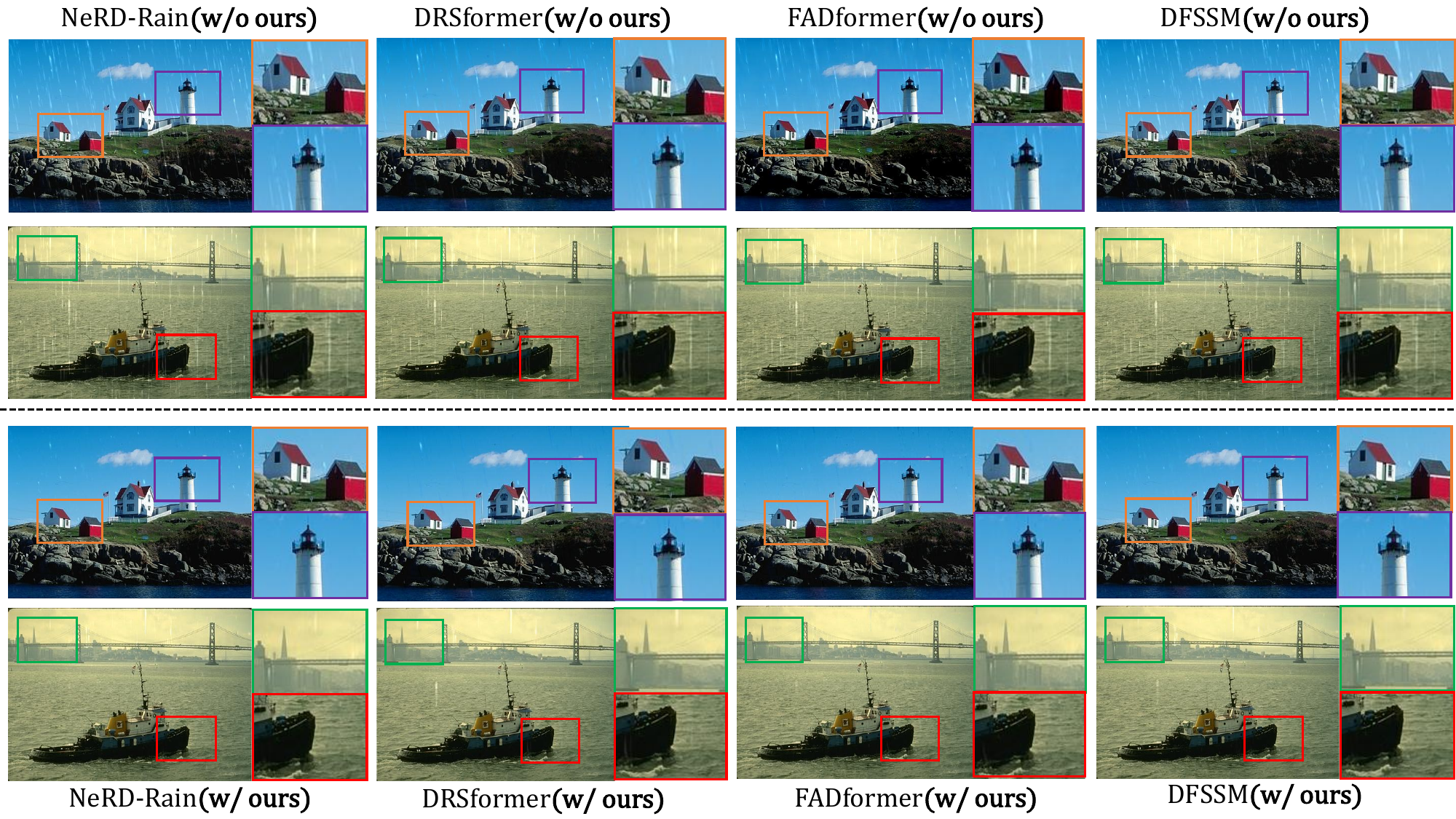}
\caption{Derained results on the Rain200L dataset. Compared with the derained results without our method, our method recovers a high-quality
image with clearer details. Zooming in the figures offers a better view at the deraining capability. }
\label{fig6}
\vspace{-3mm}
\end{figure*}

\begin{equation}
M_\theta(x, y) =
\begin{cases}
    \frac{1}{L_r}, & \text{if } (x, y) \in D_\theta, \ |(x, y)| \leq \frac{L_r}{2} \\
    0, & \text{otherwise}
\end{cases}
\end{equation}
\begin{equation}
\mathrm{I_{GM}(u,v)=\sum_{x=-L_r}^{L_r}\sum_{y=-L_r}^{L_r}I_{G}(u'-xv-y)M_{\theta}(xy)}
\end{equation}

where the rain streak length $L_r \in [L_{\min}, L_{\max}]$, inclination angle $\theta_r \in [\theta_{\min}, \theta_{\max}]$, and rain streak width $W_r \in [W_{\min}, W_{\max}]$ are all adjustable parameters. In the superposition stage, a weighted fusion strategy is adopted to superimpose rain streaks onto the luminance channel:
\begin{equation}
I_Y^{\text{rainy}} = (1-\beta) \cdot X_m + \beta \cdot I_Y
\end{equation}

where $\beta$ is the fusion coefficient. Finally, merge the processed luminance channel with the original UV channels, and convert back to RGB space, obtaining the rainy image $I_{\text{input}}^{\text{fuse}}$. This module finally outputs a set of high-correlation pseudo-paired samples ($I_{\text{input}}^{\text{fuse}}, I_{\text{target}}^{\text{fuse}}$), providing key training data for the target domain transfer of the cross-domain deraining model.

\subsection{Losses}
The proposed methodology is characterized by a plug-and-play property. In our experimental evaluations, we applied this approach to a selection of representative deraining architectures. During the training phase, the network is governed by a joint optimization strategy incorporating Charbonnier Reconstruction Loss, Frequency Domain Consistency Loss and Edge Preservation Loss. 

\emph{Charbonnier Reconstruction Loss}: This loss term serves to quantify pixel-wise discrepancies between the reconstructed derained image and the ground-truth rain-free image. Formulated as a smoothed $L_1$ penalty, it enhances the stability of convergence:
\begin{equation}
\mathcal{L}_{char} = \sqrt{(\hat{I} - I_{gt})^2 + \epsilon^2}
\end{equation}
where $\epsilon$ represents a constant typically set within the range of $10^{-3}$ to $10^{-6}$, employed to mitigate the issue of gradient discontinuity.

\emph{Frequency Domain Consistency Loss }: This loss term imposes constraints on spectral consistency between the reconstructed image and the ground-truth image within the frequency domain, placing particular emphasis on preserving textural details and structural distributions. To implement this, the Two-Dimensional Fast Fourier Transform (FFT) is first applied to the image: 
\begin{equation}
F(\hat{I}) = \text{FFT}(\hat{I}), \quad F(I_{gt}) = \text{FFT}(I_{gt})
]
\end{equation}
The frequency domain loss is defined as: 
\begin{equation}
\mathcal{L}_{fft} = \frac{1}{N} \sum_{u,v} \|F(\hat{I})_{u,v}| - |F(I_{gt})_{u,v}| |
\end{equation}

\begin{table*}[t]
    \centering
    \caption{Quantitative Comparison on OOD Dataset.}
    \label{tab:ood_comparison}
    \renewcommand{\arraystretch}{1.3} 
    \small 
    \begin{tabular}{|l|c|c|c|c|c|c|} 
        \hline 
        \multirow{2}{*}{\textbf{Baseline}} & \multicolumn{6}{c|}{\textbf{Testing Set: Rain200L (OOD)}} \\ 
        \cline{2-7} 
         & PSNR$\uparrow$ & SSIM$\uparrow$ & LPIPS$\downarrow$ & FSIM$\uparrow$ & NIQE$\downarrow$ & PI$\downarrow$ \\
        \hline \hline
        NeRD-Rain~\cite{chen2024bidirectional} w/o ours & 25.44 & 0.8372 & 0.2483 & 0.8872 & 4.1645 & 2.7664 \\ \hline
        NeRD-Rain~\cite{chen2024bidirectional} w/ ours & 33.67 \textcolor{red}{(32\%$\uparrow$)} & 0.9495 \textcolor{red}{(13\%$\uparrow$)} & 0.1000\textcolor{red}{(60\%$\uparrow$)} & 0.9559\textcolor{red}{(8\%$\uparrow$)} & 3.2822\textcolor{red}{(21\%$\uparrow$)} & 2.3327\textcolor{red}{(16\%$\uparrow$)} \\ \hline \hline
        DRSformer~\cite{chen2023learning} w/o ours & 28.35 & 0.8787 & 0.1969 & 0.9068 & 3.6658 & 2.5371 \\ \hline
        DRSformer~\cite{chen2023learning} w/ ours & 32.68 \textcolor{red}{(15\%$\uparrow$)} & 0.9478 \textcolor{red}{(7\%$\uparrow$)} & 0.0849\textcolor{red}{(57\%$\uparrow$)} & 0.9547\textcolor{red}{(5\%$\uparrow$)} & 3.2469\textcolor{red}{(11\%$\uparrow$)} & 2.3505\textcolor{red}{(11\%$\uparrow$)} \\ \hline \hline
        FADformer~\cite{gao2024efficient} w/o ours & 27.18 & 0.8535 & 0.2375 & 0.8915 & 4.1380 & 2.7486 \\ \hline
        FADformer~\cite{gao2024efficient} w/ ours & 33.85 \textcolor{red}{(24\%$\uparrow$)} & 0.9550 \textcolor{red}{(11\%$\uparrow$)} & 0.0747\textcolor{red}{(69\%$\uparrow$)} & 0.9609\textcolor{red}{(8\%$\uparrow$)} & 3.2139\textcolor{red}{(22\%$\uparrow$)} & 2.3227\textcolor{red}{(15\%$\uparrow$)} \\ \hline \hline
        DFSSM~\cite{yamashita2024image} w/o ours & 27.09 & 0.8499 & 0.2402 & 0.8879 & 4.1527 & 2.8573 \\ \hline
        DFSSM~\cite{yamashita2024image} w/ ours & 33.30 \textcolor{red}{(22\%$\uparrow$)} & 0.9490 \textcolor{red}{(11\%$\uparrow$)} & 0.0804\textcolor{red}{(67\%$\uparrow$)} & 0.9589\textcolor{red}{(8\%$\uparrow$)} & 3.2245\textcolor{red}{(22\%$\uparrow$)} & 2.3367\textcolor{red}{(18\%$\uparrow$)} \\ 
        \hline 
    \end{tabular}
    \label{tab1}
\end{table*}

\begin{table}[!ht]
    \small
    \centering
    \caption{Comparison of Training Efficiency and Performance}
    \label{tab:efficiency_comparison}
    \renewcommand{\arraystretch}{1.2}
    \begin{tabular}{l|l|c|c|c}
        \hline
        \textbf{Baseline} & \textbf{Variant} & \textbf{Best Ep.} & \textbf{PSNR$\uparrow$} & \textbf{SSIM$\uparrow$} \\ 
        \hline \hline
        \multirow{2}{*}{NeRD-Rain} & random & 84 & 27.70 & 0.8779 \\ \cline{2-5}
         & \textbf{Ours} & \textbf{15 (\textcolor{red}{5.6$\times$})} & \textbf{33.67} & \textbf{0.9495} \\ \hline
        \multirow{2}{*}{DRSformer} & random & 26 & 24.52 & 0.8515 \\ \cline{2-5}
         & \textbf{Ours} & \textbf{6 (\textcolor{red}{4.3$\times$})} & \textbf{32.68} & \textbf{0.9478} \\ \hline
        \multirow{2}{*}{FADformer} & random & 150 & 28.70 & 0.8894 \\ \cline{2-5}
         & \textbf{Ours} & \textbf{101 (\textcolor{red}{1.5$\times$})} & \textbf{33.85} & \textbf{0.9550} \\ \hline
    \end{tabular}
    \label{tab2}
\end{table}

\emph{Edge Preservation Loss:} This loss term enhances the consistency of structural edges by imposing constraints on image gradients. It specifically utilizes Sobel or Laplacian operators to compute the edge maps:
\begin{equation}
E(\hat{I}) = \nabla(\hat{I}), \quad E(I_{gt}) = \nabla(I_{gt})
\end{equation}
The loss function is defined as follows: 
\begin{equation}
\mathcal{L}_{edge} = | E(\hat{I}) - E(I_{gt}) |_1
\end{equation}
where $\nabla$ denotes the gradient operator (specifically, the Sobel convolution kernel).
Finally,the aggregate objective function is formulated as follows: 
\begin{equation}
\mathcal{L}{total} = \lambda_{1} \mathcal{L}_{char} + \lambda_{2} \mathcal{L}_{fft} + \lambda_{3} \mathcal{L}_{edge} 
\end{equation}
where $\mathcal{L}_{char}$ denotes the Charbonnier Reconstruction Loss (facilitating stable convergence within the pixel domain), $\mathcal{L}_{fft}$ represents the Frequency Domain Consistency Loss (constraining high-frequency details), and $\mathcal{L}_{edge}$ signifies the Edge Preservation Loss (enhancing structural restoration).

\begin{figure*}[ht]
\centering
\includegraphics[width=0.95\linewidth]{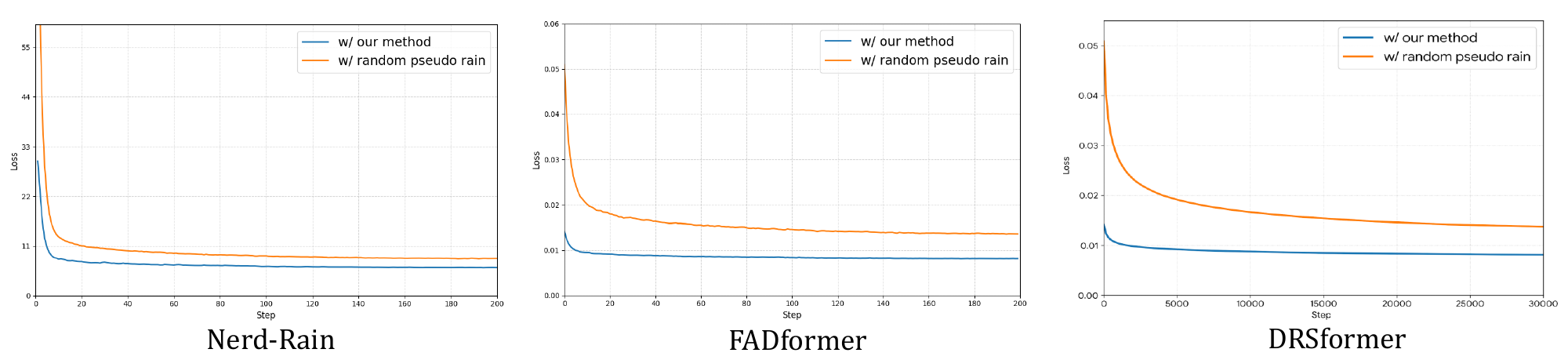}
\caption{The ablation study of our algorithm on different baseline methods. Our algorithm has greatly enhanced the \textbf{convergence speed and generalization capability} on cross-scenario environments. }
\label{fig7}
\vspace{-3mm}
\end{figure*}

\section{Experimental Validation}
\subsection{Experimental Setup}

\emph{Implementation Details}: The proposed framework is implemented using PyTorch, with all experiments conducted on an NVIDIA GeForce RTX 3090 GPU. During the training phase, the patch size is configured to $128 \times 128$ and the batch size to 12. The learning rate is initialized at $5 \times 10^{-5}$and subsequently decayed to $5 \times 10^{-7}$ utilizing a cosine annealing strategy over 200 epochs.

\emph{Out-of-Distribution Derain Benchmark}: 
To simulate Out-of-Distribution (OOD) scenarios, we utilize HQ-RAIN~\cite{chen2025towards} as the source dataset and Rain200L~\cite{yang2017deep} as the target dataset. Specifically, the source dataset comprises paired samples with and without rain (w/ rain and w/o rain), whereas the target dataset contains exclusively clean, rain-free images (w/o rain)

\subsection{Experimental Results}

On the Out-of-Distribution Derain Benchmark, we integrated our approach into four representative methods: NeRD-Rain~\cite{chen2024bidirectional}, FADformer~\cite{gao2024efficient}, DRSformer~\cite{chen2023learning}, and DFSSM~\cite{yamashita2024image}. For a holistic assessment of deraining efficacy across both pixel-level fidelity and perceptual quality, we expand our quantitative framework to encompass a bifurcated suite of metrics: full-reference (PSNR, SSIM, LPIPS, and FSIM) and no-reference (NIQE and PI) indicators.   

Table.~\ref{tab1} presents the experimental results for Out-of-Distribution (OOD) scenarios. It is observed that the three model frameworks, when trained originally on the source domain (independent identically distributed, IID) dataset, yield suboptimal results when evaluated on the target domain (OOD) test set. To address this, we utilize the source-domain model as a pre-trained baseline and conduct transfer training leveraging the newly synthesized pseudo-data. Experimental results demonstrate that our approach yields a substantial gain in PSNR (18\% to 35\%) and SSIM (8\% to 15\%) across various benchmarks. The marked optimization in perceptual metrics quantitatively substantiates the efficacy of the Super-pixel Generation (Sup-Gen) module; specifically, the dramatic reduction in LPIPS  alongside a peak FSIM of 0.9625 underscores the module's capability in preserving intrinsic structural priors and intricate textures within the target domain. Additionally, the no-reference indicators NIQE and PI as proxies for image naturalness exhibit significant declines of 2\% and 20\% respectively to validate the superior fidelity of the restored images. Comparisons in Fig.~\ref{fig6} illustrate the significant advantages of our method in image deraining tasks. Specifically, in the regions highlighted by green and red boxes, the results demonstrate superior preservation of background content and textural details. Existing benchmark models frequently suffer from persistent rain streak artifacts or induce undesirable blurring of the underlying background details. Consequently, the proposed method exhibits remarkable adaptability to the target domain. Our approach reconstructs images characterized by aesthetic appeal, high naturalness, and sharp edge definitions, which is in seamless alignment with the superior FSIM and NIQE metrics achieved.

\begin{table}[t]
    \centering
    \caption{Ablation Study on Resolution-adaptive Fusion Scale ($\alpha$)}
    \label{tab: ablation_alpha}
    \renewcommand{\arraystretch}{1.2}
    \begin{tabular}{|l|c|c|c|c|}
        \hline
        \boldmath$\alpha$ & 0.8 & 0.6 & 0.4 & \textbf{0.2} \\
        \hline
        PSNR $\uparrow$ & 32.84 & 32.83 & 32.59 & \textcolor{red}{33.08} \\
        \hline
        SSIM $\uparrow$ & 0.9427 & 0.9420 & 0.9412 & \textcolor{red}{0.9460} \\
        \hline
    \end{tabular}
\label{tab4}
\end{table}

\subsection{Comparison with Random Rain Streaks}

As shown in Table.~\ref{tab2}, "w/ random pseudo rain" denoted as simple random rain streaks are directly superimposed onto the target domain images. We conducted comparative experiments benchmarking this simple rain generation baseline against our proposed method under identical environmental conditions. Although the random rain streak benchmark achieves limited OOD adaptation, it fails to capture the physical dynamics and optical gradients of real-world rain, leading to poor perceptual quality. In contrast, our multi-stage Pseudo-label Re-Synthesis (Pseudo-Syn) module provides physics-inspired high-fidelity pseudo-paired samples, significantly accelerating the convergence of various architectures. Experimental results confirm that the proposed framework not only substantially improves training efficiency but also ensures robust and perceptually natural image restoration under severe domain shifts. 

\subsection{Training Efficiency Comparison}
As shown in Table.~\ref{tab2}, in the three sets of efficiency comparison experiments, we recorded the trajectories of the loss curves for each group, as illustrated in Fig.~\ref{fig7}. In terms of intra-group comparison, our method demonstrates faster convergence to the optimal state. Regarding random group comparison, our method maintains a lower average loss compared to the random group baselines, indicating superior robustness across different experimental settings.

\subsection{Ablations}

\textbf{Ablation on Superpixel Generation:} To evaluate the effectiveness of the Sup-Gen module, specific ablation studies were conducted. Initially, we established a baseline simulating the absence of our method (denoted as "w/o ours method") by the direct superposition of random image patches. Subsequently, we investigated an alternative configuration by substituting the SLIC~\cite{6205760} algorithm with the NC05~\cite{cour2005spectral} algorithm. Quantitatively, the NC05 configuration yields a PSNR of 32.56 dB, compared to 32.96 dB for the 'w/o ours' variant. Benchmarking these variations against our proposed approach which yields a PSNR of 33.35db revealed that our method yields substantial improvements compared to the "w/o ours method" baseline. Conversely, the utilization of the NC05~\cite{cour2005spectral} algorithm resulted in observable performance degradation within the deraining task.

\textbf{Ablation on Resolution-adaptive Fusion Rate $\alpha$:} As Table.~\ref{tab4} shows, this table investigates the influence of the fusion coefficient $\alpha$ within the Resolution-Adaptive Fusion Equation. \eqref{my_formula}. We synthesized four distinct sets of pseudo-data corresponding to $\alpha$ values of 0.8, 0.6, 0.4, and 0.2, while maintaining all other variables constant, and subsequently performed transfer training. Empirical observations indicate that the optimal transfer performance is achieved when $\alpha$ is set to 0.2.


\section{Conclusion}
This paper presents a cross-scenario image deraining adaptation methodology anchored in superpixel partitioning and pseudo-rain streak synthesis. By leveraging superpixel patch generation and fusion modules, the framework effectively incorporates structural priors from the source domain. Furthermore, by integrating a multi-stage rain streak synthesis strategy, it facilitates the generation of high-fidelity pseudo-labeled data. Exhibiting robust generalization and portability, this approach offers an effective solution for image deraining in real-world applications. 

\bibliographystyle{IEEEtran}
\bibliography{ref}

\begin{IEEEbiography}
[{\includegraphics[width=1in,height=1.25in,clip,keepaspectratio]{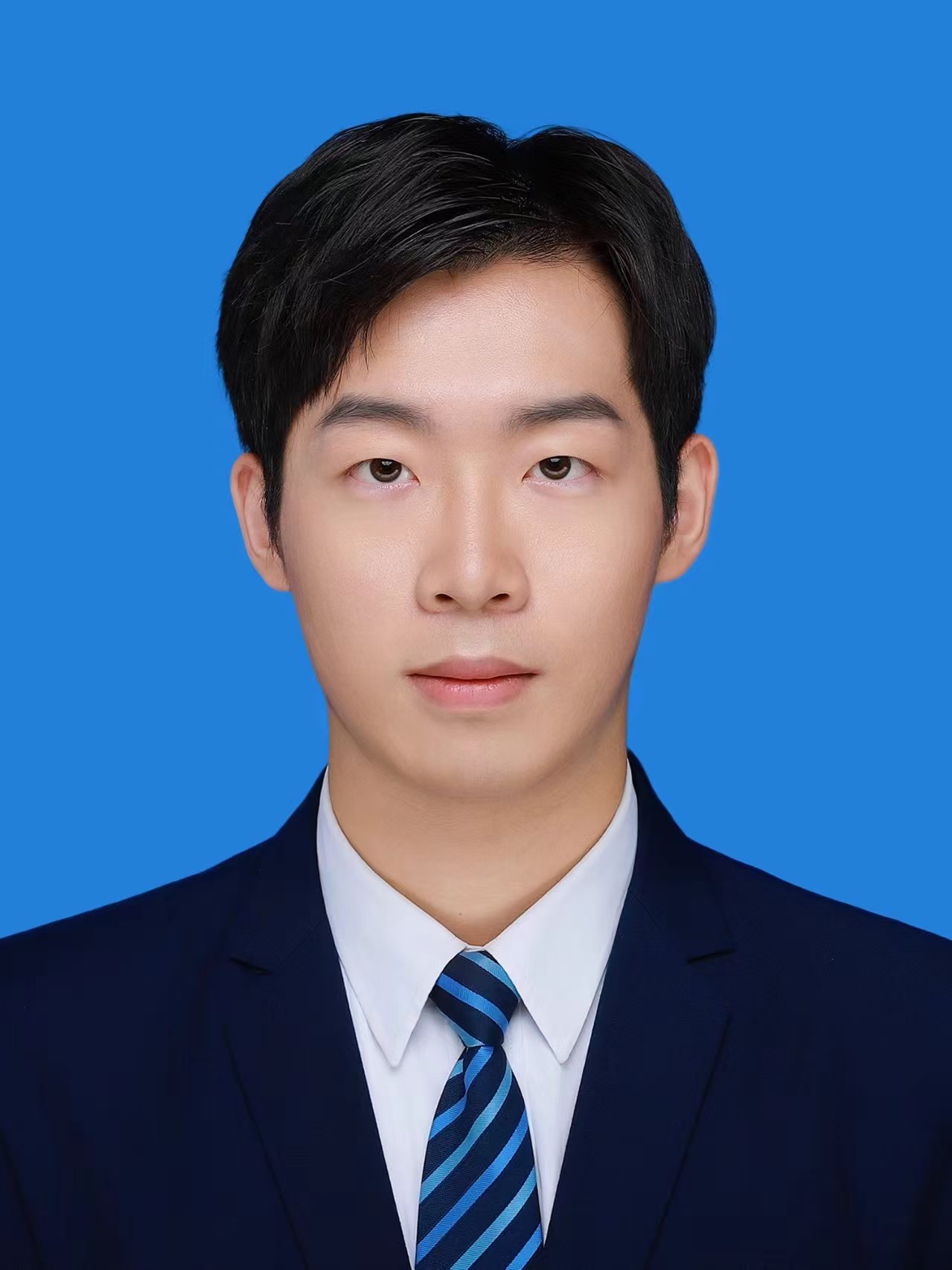}}]{Kangbo Zhao } received the B.S. degree in 2024, from the School of Information Engineering, Guangdong University of Technology, Guangzhou, China, where he is currently working towards a M.S. degree. His research interests include computer vision and machine learning. 
\end{IEEEbiography}

\begin{IEEEbiography}[{\includegraphics[width=1in,height=1.25in,clip,keepaspectratio]{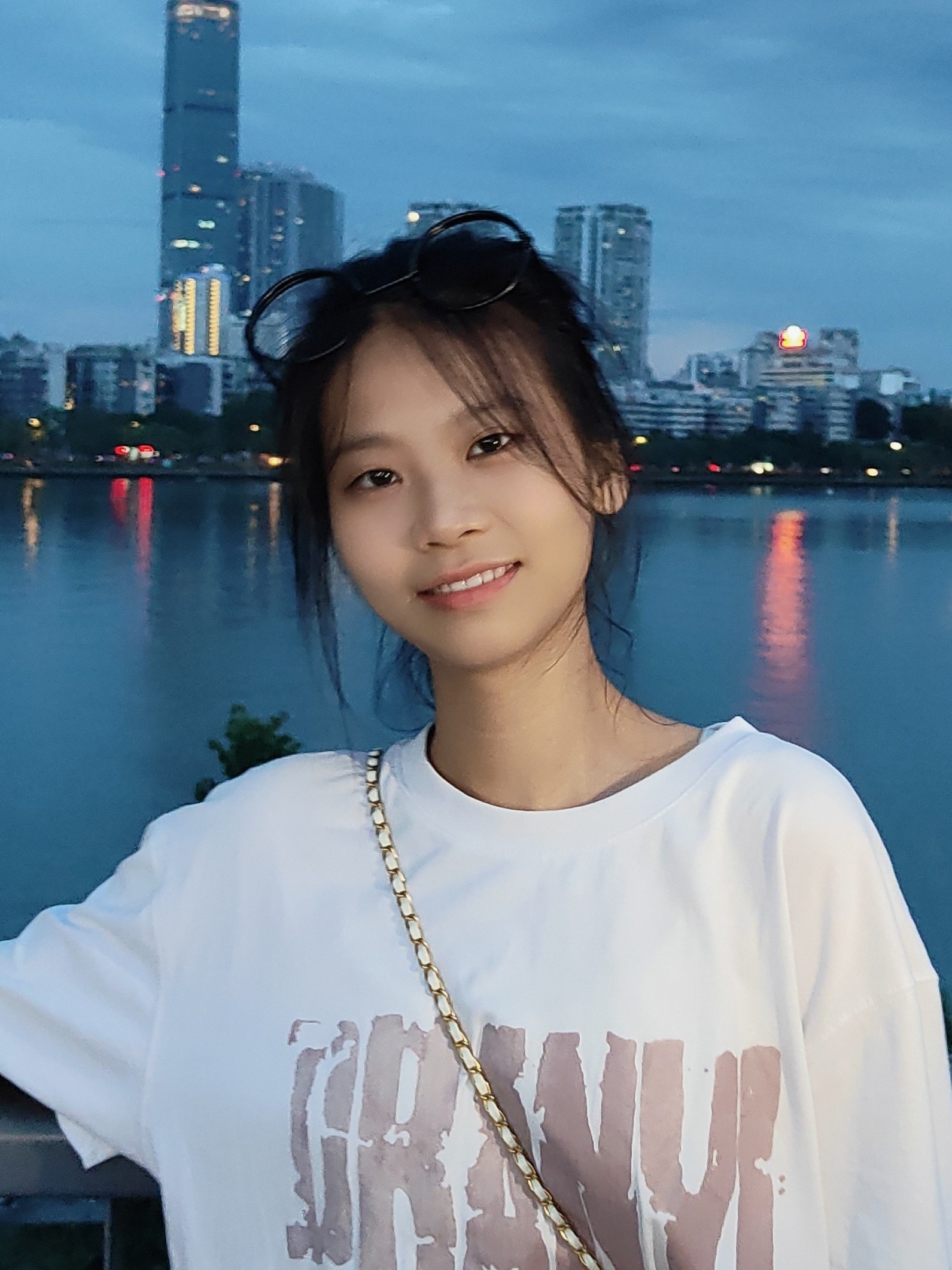}}]{Miaoxin Guan }received the B.S. degree from the School of Electronic Engineering, South China Agricultural University, Guangzhou, China, in 2025. She is currently working towards the M.S. degree at the School of Information Engineering, Guangdong University of Technology, Guangzhou, China. Her research interests include computer vision and machine learning.
\end{IEEEbiography}

\begin{IEEEbiography}
[{\includegraphics[width=1in,height=1.25in,clip,keepaspectratio]{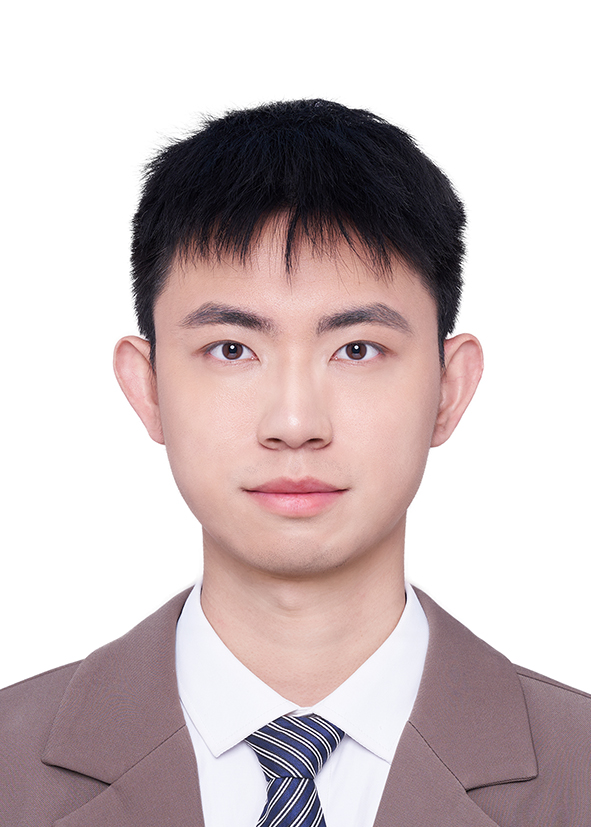}}]{Xiang Chen}
received the MS degree from the College of Electronic and Information Engineering, Shenyang Aerospace University, China, in 2022. He is currently working toward the PhD degree with the School of Computer Science and Engineering, Nanjing University of Science and Technology, China. His research interests include computer vision and deep learning, with special emphasis on image restoration and enhancement under adverse weather conditions.
\end{IEEEbiography}

\begin{IEEEbiography}
[{\includegraphics[width=1in,height=1.25in,clip,keepaspectratio]{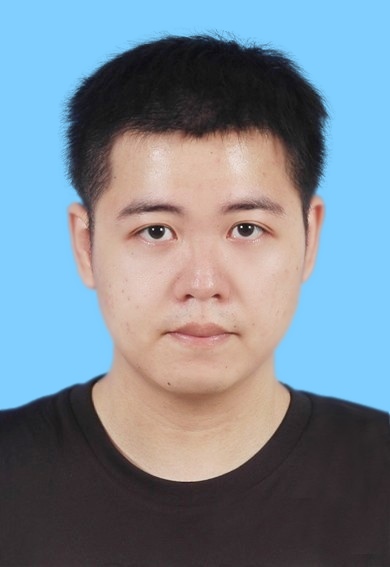}}]{Yukai Shi}
received the Ph.D. degree from the School of Data and Computer Science, Sun Yat-sen University, Guangzhou China, in 2019. He is currently an associate professor at the School of Information Engineering, Guangdong University of Technology, China. His research interests include computer vision and machine learning.
\end{IEEEbiography}

\begin{IEEEbiography}
[{\includegraphics[width=1in,height=1.25in,clip,keepaspectratio]{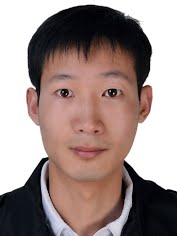}}]{Jinshan Pan}(Senior Member, IEEE) received the PhD degree in computational mathematics from the Dalian University of Technology, China, in 2017. He was a joint-training PhD student with the School of Mathematical Sciences, Dalian University of Technology and also with Electrical Engineering and Computer Science, University of California, Merced, CA. He is currently a professor with the School of Computer Science and Engineering, Nanjing University of Science and Technology. His research interests include image deblurring, image/video analysis and enhancement, and related vision problems. He is a senior member of IEEE.
\end{IEEEbiography}

\end{document}